\documentclass[runningheads]{llncs}

\usepackage{graphicx}
\usepackage{multirow}
\usepackage{booktabs}
\usepackage{cite}
\usepackage{xcolor}
\usepackage{url}
\usepackage[normalem]{ulem}

\begin{document}
\title{\emph{ReAct}: A \emph{Re}view Comment Dataset for \emph{Act}ionability (and more)}

\author{Gautam Choudhary\inst{1} \and Natwar Modani\inst{1} \and Nitish Maurya\inst{2}}
\authorrunning{Gautam Choudhary, Natwar Modani, Nitish Maurya}

\institute{Adobe Research \and Adobe System \\
\email{\{gautamc, nmodani, nmaurya\}@adobe.com}}

\maketitle              

\begin{abstract}
Review comments play an important role in the evolution of documents. 
For a large document, the number of review comments may become large, making it difficult for the authors to quickly grasp what the comments are about. 
It is important to identify the nature of the comments to identify which comments require some action on the part of document authors, along with identifying the types of these comments. 
In this paper, we introduce an annotated review comment dataset \emph{ReAct}. 
The review comments are sourced from OpenReview site. 
We crowd-source annotations for these reviews for \emph{actionability} and \emph{type} of comments.
We analyze the properties of the dataset and validate the quality of annotations.
We release the dataset to the research community as a major contribution\footnote{Full dataset available at \url{https://github.com/gtmdotme/ReAct}}.
We also benchmark our data with standard baselines for classification tasks and analyze their performance. 
\end{abstract}

\section{Introduction}

Review comments play an important role in the evolution of documents. Academic publications routinely go through a peer-review process, where the reviewers provide both their opinion about the suitability of the articles for the publication venue and also feedback to the authors for potentially improving the contributed article. Further, several publication venues are providing the authors a chance to respond to the review comments (for example, ACL, NAACL, EMNLP, etc., in addition to most journals). Therefore, it is important for the authors to be able to quickly digest the review comments so that they can address the concerns of the reviewers and clarify certain points which may not have been communicated adequately by the article itself.

In this work, we focus on two aspects of \textit{understanding} the review comments. First, determining if a review comment requires some action on the part of document authors. This motivates the need for the classification of review comments based on `actionability'. Second, what type of review comment it is among \textit{Agreement, Disagreement, Question, Suggestion, Shortcoming, Statement of Fact}, and \textit{Others}, similar to (but not exactly the same as)~\cite{sancheti2019understanding}. 
We provide the reason for our choice of these specific types and their justification in Section~\ref{sec:classChoice}. 

Text classification has long been an active area of research, as the classification can help the users efficiently process a large amount of content. Finding actionable comments on social media (tweets) was addressed in~\cite{spasojevic2015identifying} using new lexicon features. A \textit{specificity} score was explored in~\cite{deshpande2010unsupervised} for an employee satisfaction survey and product review settings to understand actionable suggestions and grievances (complaints) for improvements. 
In yet another work~\cite{meyers2018dataset}, the actionability of review comments for code review is investigated using lexical features.
These works address only the actionability aspect of our problem, and the datasets used in these papers are not publicly available except in~\cite{meyers2018dataset}, where the dataset is made available publicly. 

Other binary classifications in prior work include Question classification~\cite{zhang2003question}, agreement/disagreement classification~\cite{ahmadalinezhad2018detecting} and suggestions/advice mining~\cite{dong2013automated}. However, such binary classifications only provide information on a single dimension in isolation and fall short in providing a more extensive set of categorization as done in \cite{mukherjee2012modeling}, where the authors investigated comments on product reviews in an e-commerce setting. 
Again, the datasets are not publicly available, and the categories proposed are not comprehensive in nature.

OpenReview is a popular online forum for reviewing research papers and the choice of gathering data from this forum is motivated by a comprehensive study for analyzing the review process~\cite{tran2020open}.
~\cite{kang2018dataset} also present \textit{PeerRead} dataset consolidating reviews from a lot of conferences. 
Our dataset provides finer-grained annotation by providing two labels per review comment sentence and thereby opens up a new research direction. 

Our key contributions in this paper are:
\begin{itemize}
    \item A review comment dataset consisting of $1,250$ labeled comments for identifying actionability and their types.
    We also have $\sim 52k+$ unlabelled (but otherwise processed) comments in this dataset for future extensions and/or use of semi-supervised approaches.
    \item A taxonomy for types of review comments.
    \item Establishing strong baselines for the proposed dataset.
\end{itemize}
\section{Dataset: ReAct}
\label{sec:data}
While the prior art focuses on feature engineering and model architecture, we note a lack of publicly available datasets in this problem set. This section describes how we arrive at the proposed annotated dataset, \emph{ReAct}.

In this paper, We use Fleiss' kappa $\kappa$~\cite{fleiss1971measuring} as the measure of inter-annotator agreement. It is used to determine the level of agreement between two or more annotators when the response variable is measured on a categorical scale. 

\subsection{Raw Data Collection and Preprocessing}
The proposed dataset is gathered from an online public forum OpenReview where research papers are reviewed and discussed. Multiple anonymous reviewers review the papers and write free-form comments (along with people other than reviewers also writing comments) related to the papers. 
We extract $911$ papers submitted to ICLR (The International Conference on Learning Representations) 2018 from OpenReview and filter out the comments written by people other than the reviewers and get a dataset where each record is a paper associated with its reviews and other metadata (final decision, rating, link to the paper, timestamps, abstract, etc.). 
Each paper is reviewed by at least $3$ reviewers who provide their comments in free-form text. An average review spans about $19$ sentences. 
On manual inspection, we find that giving a specific label to the whole review is not appropriate, since a review often contains some facts about the paper along with some merits/demerits and some questions. A natural choice is to chunk the review into smaller units. 
We follow a simple approach and choose a sentence as the atomic unit and refer to it as a \textit{comment}.
We use a python tool pySBD\footnote{pySBD: \url{https://github.com/nipunsadvilkar/pySBD}} for splitting and disambiguation of long paragraphs of reviews into sentences.

\subsection{Classification Taxonomy}
\label{sec:classChoice}
Given the motivating scenario of helping the user quickly be able to respond to review comments, say during rebuttal period, the choice of binary classification (as \textit{actionable} or not) is fairly straightforward and has been used in prior literature as already discussed (although, not in paper review setting). However, the choice of \textit{types} for the finer-grained classification is non-obvious.

To arrive at the appropriate class labels, we randomly selected $50$ review comments and three volunteers started categorizing them independently, with an initial \textit{types} seed list of \textit{Suggestion}, \textit{Agreement}, \textit{Disagreement}, and \textit{Question}, inspired by~\cite{sancheti2019understanding}. Whenever the volunteers felt that the comment didn't fall in these types, they added a new type. After completing the independent categorization, a pool of all labels for types across the three volunteers was created. Now, a set of labels was consolidated if the volunteers agreed that the individual labels in that set had the same semantics. 
The final proposed taxonomy of category labels with their initial label assignments in parenthesis were found to be as follows: agreement (appreciation/agreement), disagreement (conflict/disagreement), question (inquiry/question), suggestion (demand/ask/advice/suggestion), shortcoming (problem/issue/shortcoming), fact (opinion/statement of fact), and others (miscellaneous/others).

\subsection{Designing the Survey}
We selected $125$ reviews for the main survey such that they were sufficiently long for having (at least) $10$ comments as part of each review, and retained $10$ randomly selected sufficiently long comments (having at least 10 words) corresponding to each of these reviews.
These $1250$ review comments are then annotated by a popular crowdsourcing platform, Amazon Mechanical Turk (AMT)\footnote{Amazon Mechanical Turk: \url{https://mturk.com/}}. Each comment is annotated by $5$ different human annotators.
The annotators are given a set of instructions for annotating the comments.
The task of an annotator is to read the review comments and assign two labels to each comment. 
The first label $Label_1$ is to be assigned based on the actionability of the comment, i.e., among \{\textit{yes}, \textit{no}\} and constitutes $Task_1$. Similarly, $Label_2$ is to be assigned from the proposed taxonomy, one of \{\textit{agreement}, \textit{disagreement}, \textit{suggestion}, \textit{question}, \textit{fact}, \textit{shortcoming}, \textit{other}\} based on the type of comments and constitutes $Task_2$. 
We also asked annotators to provide feedback for the survey to mitigate shortcomings, if any.
The survey was available to annotators based on certain filters that AMT provides. We restricted the survey to \textit{Mechanical Turk Masters} who had acceptance scores $\geq 95 \%$ to get high-quality annotations. 
The reward of one complete survey (comprising two types of labels for $10$ comments) was set to \$ $0.75$ based on the feedback received on the pilot surveys floated initially, described next. A time limit of $30$ minutes was set before the survey expired.

\subsection{Analyzing Responses}

\subsubsection{Pilot Survey}
\label{sec:pilot}
Instead of rolling out the survey fully in one go, we followed an iterative approach. A pilot survey was conducted to check if the tasks and instructions were clear and to get an estimate of the quality of responses. We handpicked $5$ reviews (different from the ones used in the main survey) having a total of $50$ comments using the above survey design. 
Post completion, we analyzed each of the responses one by one and noted a `moderate' inter-annotator agreement score (Fleiss kappa), $\kappa  \approx 0.48$ among the annotators~\cite{landis1977measurement}.
A noteworthy thing was the feedback received from annotators which substantially supported our comprehensive, yet simple survey design. The annotators expressed no lack of clarity and also found the survey task to be appropriate and complete.
The time to answer was also analyzed and seemed to match with the time taken by the volunteers (close to $10$ minutes per survey). Hence, the feedback received from this survey was sufficiently positive to go ahead with the main survey.

\subsubsection{Main Survey}
Post successful completion of the pilot survey, we obtain a set of $6,250$ annotations for the above chosen 1250 comments (each labeled by $5$ different annotators). 
Each annotation consists of labels for the comment along with other metadata such as characteristics of the annotator and the survey such as (ID, duration, timestamp, etc.). 
We found that a total of $33$ unique annotators participated in the survey with an average 
completion time of $\sim 10$ minutes.

\begin{figure}[!bt]
	\centering
	\includegraphics[width=\linewidth]{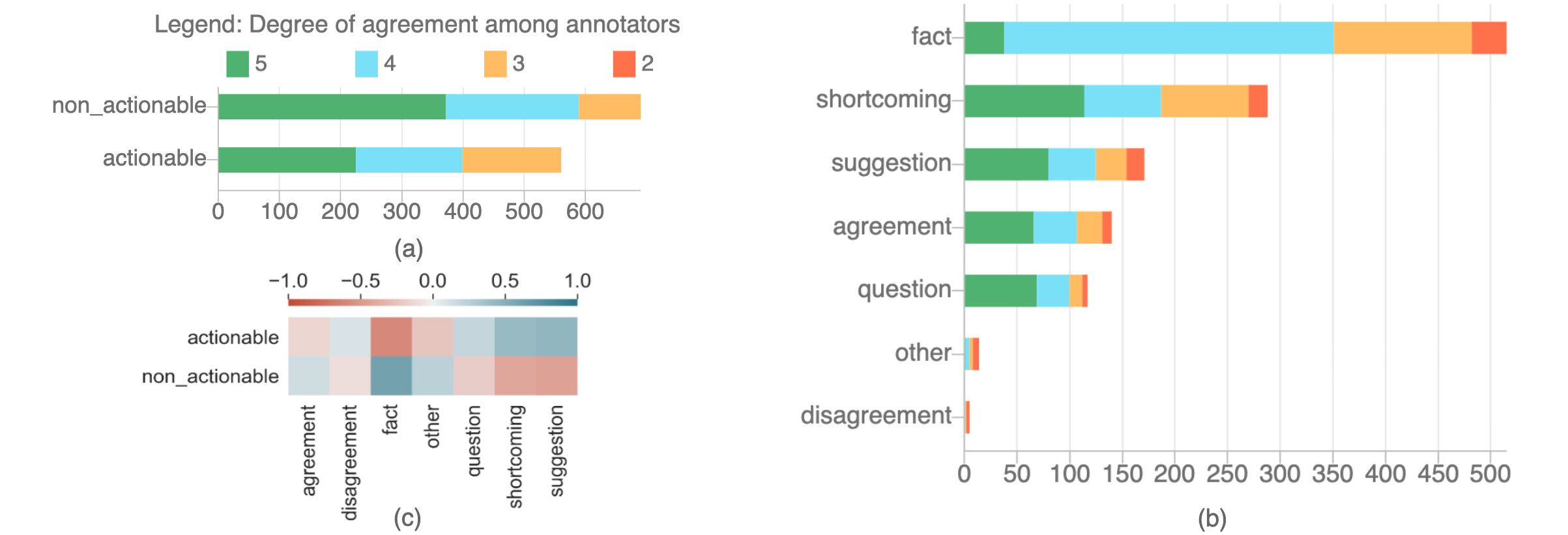}
    \caption{(a,b) Distribution of review comments based on their count, hued by the fraction of agreement in annotators annotating the same comment for $Label_1$ and $Label_2$ respectively, (c) correlation observed in the two kinds of category labels annotations.}
	\label{fig:label-consensus}
\end{figure}

First, we analyze the Fleiss kappa scores on individual labeling tasks, i.e., for $Label_1$ and $Label_2$. For the $Label_1$ denoting actionability, we observe a `moderate' inter-annotator score of $0.49$ and a slightly higher score of $0.53$ for $Label_2$ based on the proposed taxonomy~\cite{landis1977measurement}. Next, we analyze at a deeper level by looking at proportions of responses ranging from having a clear agreement to strong ambiguity as shown by proportions of stacked bars in Figure~\ref{fig:label-consensus}(a) and (b). At an aggregate level, for $Label_1$, almost $50 \%$ of annotations have a clear consensus where all $5$ annotators vote for the same category label, while $30 \%$ of annotations have $4$ out of $5$ votes and the rest $20 \%$ have $3$ out of $5$ votes as shown in Figure~\ref{fig:label-consensus}(a) at per category label basis. Similarly, for $Label_2$, more than $70 \%$ of annotations have at least $4$ out of $5$ annotators agree on a specific category label as shown in Figure~\ref{fig:label-consensus}(b). We observe that \textit{disagreement} is a rare class 
(with a low agreement between reviewers), suggesting this label may not be essential.

Additionally, the correlation analysis using the Pearson Correlation Coefficient between the two sets of category labels (Figure~\ref{fig:label-consensus}(c)) strengthens the intuition that suggestions, shortcomings, and questions are more likely to be actionable than the other categories. 

\subsection{Moderation}
\label{sec:moderation}

We found some noisy responses where annotators labeled shortcomings as non-actionable. Another example of noise found in the data is when annotators annotate an agreement as actionable. The proportion of such noisy responses was very small ($\sim 6.5\%$) in the whole dataset. To improve the quality of data further, we selectively moderated the labeling. In particular, we reviewed $Label_2$ for review comments in the case of maximum disagreement, i.e., when the maximum number of annotators agreeing on that label was $2$. In addition, if the maximum number of annotators agreeing on $Label_1$ was $3$ for any of \textit{these} comments, then this label was also reviewed. In the cases where the maximum number of annotators agreeing $Label_2$ was $3$ (or more), indicating a good level of agreement ($3$ out of $5$), neither of the labels were reviewed (even if for $Label_1$, the maximum number of agreeing annotators were $3$, the borderline case). 

There were a total of $91$ cases where we reviewed $Label_2$, out of which $49$ cases where we also reviewed $Label_1$. Ignoring these cases, the inter-annotator agreement for $Label_1$ increased to $0.52$, and for $Label_2$, it increased to $0.57$.
Finally, in review, we ended up changing $Label_1$ for $19$ cases.
For $Label_2$, there were $34$ cases where there was a tie and we picked one of the tied labels. Further, we changed labels for $18$ other cases, where we assigned a label as ground truth for $Label_2$, which was not voted as (one of) majority label(s) by annotators.
Given the small number of cases where we had to change the labels (about $1.5\%$ cases), we believe the annotations' quality is very good.

\subsection{Processed Dataset}
Based on the above analysis, we assign the ground truth labels based on majority, i.e., out of $5$ votes for a given comment, we chose the majority vote as the ground truth label. While a tie is not possible in $Task_1$, it may take place in $Task_2$ (consider the case where all the votes are for different labels or two votes each for two labels). Total number of tied cases for $label_2$ was $34$. As mentioned before, such ties were resolved through the process of moderation.
The final prepared dataset (also available at \url{https://github.com/gtmdotme/ReAct}) consists of $1,250$ comments with two sets of labels, $Label_1$ and $Label_2$. For example, ``\textit{It would enhance readability of the paper if the results were more self-contained}'' is labeled as actionable and suggestion. A contrasting example of non-actionable and agreement labels is, ``\textit{Indeed, the authors have succeed in showing that this is not necessarily the case}''.
\section{Benchmarking Experiments}
\label{sec:method}
Given a review comment from the proposed dataset, two classification scenarios arise. First, a binary classification ($Task_1$) to identify whether the comment is actionable to the author or not, and secondly, a multiclass classification ($Task_2$) to identify the nature of comment from the proposed taxonomy.

\subsection{Feature Extraction and Text Classification Models}
Most of the recent works on producing contextual embeddings have been shown to improve results over the human crafted features. Therefore, we experiment with the following state-of-the-art sentence embeddings:

\begin{itemize}
    \item Universal Sentence Encoder~\cite{cer2018universal} (\textbf{USE}) is trained and optimized for text, such as sentences, phrases, or short paragraphs (embedding length $ = 512$).
    \item \textbf{DistilBERT} Embeddings~\cite{sanh2019distilbert} are a distilled version of BERT with faster performance and fewer parameters (embedding length $ = 768$).
    \item \textbf{RoBERTa} Embeddings~\cite{liu2019roberta} are built out of tweaking the BERT model hyperparameters to produce robust embeddings that are shown to perform best for STS tasks (embedding length $ = 1024$). 
\end{itemize}
We experiment with the following text classifiers:
\begin{itemize}
    \item \textit{Baseline-Random}: This model predicts a class uniformly at random (out of 2 for $Task_1$ and out of 7 for $Task_2$).
    \item \textit{Baseline-Majority}: This model predicts the most frequently occurring class in the training dataset.
    \item ML-Classifiers: We use standard classification models like Logistic Regression (\textit{LR}), Support Vector Machine (\textit{SVM}), \textit{XGBoost}, and Feedforward Neural Network (\textit{FNN}) with two hidden layers of sizes $128$ and $32$.
\end{itemize}

\subsection{Experimental Setup}

All the experiments involve using the proposed dataset by keeping 80\% of the data for training and the rest 20\% for evaluating the model. The (random) split is such that the proportions of the $Label_2$ are preserved in the two sets, i.e., stratified split. We use a standard implementation of these models from the scikit-learn python library\footnote{scikit-learn: \url{https://scikit-learn.org/}} keeping the default parameters fixed for a fair comparison across variations in models and embeddings.

\begin{table*}[!bt]
    \caption{Test accuracy and F1 scores of the classification models for each task on our proposed dataset. Here, DistB is DistilBERT and RoB is RoBERTa.}
    \centering
    \begin{tabular}{|r|c|c|c|c|c|c|c|c|c|c|c|c|}
    \hline
    \multirow{3}{*}{\textbf{Models}} & \multicolumn{6}{c|}{\textbf{$Task_1$}}       & \multicolumn{6}{c|}{\textbf{$Task_2$}}       \\ \cline{2-13} 
                                     & \multicolumn{3}{c|}{\textbf{Accuracy}}   & \multicolumn{3}{c|}{\textbf{F1-Score}}   & \multicolumn{3}{c|}{\textbf{Accuracy}}   & \multicolumn{3}{c|}{\textbf{F1-Score}}   \\ \cline{2-13}
                                     & \textbf{USE} & \textbf{DistB} & \textbf{RoB} & \textbf{USE} & \textbf{DistB} & \textbf{RoB} & \textbf{USE} & \textbf{DistB} & \textbf{RoB} & \textbf{USE} & \textbf{DistB} & \textbf{RoB} \\
                                     \hline
    \textit{Baseline-Random}         & 0.504 & 0.528 & 0.500 & 0.551 & 0.487 & 0.480 & 0.132 & 0.116 & 0.136 & 0.164 & 0.180 & 0.159  \\ 
    \textit{Baseline-Majority}       & 0.588 & 0.588 & 0.588 & 0.435 & 0.435 & 0.435 & 0.408 & 0.408 & 0.408 & 0.236 & 0.236 & 0.236   \\ 
    \hline
    \textit{LR} & 0.788 & 0.812 & 0.812 & 0.788 & 0.813 & 0.813 & 0.616 & 0.688 & 0.688 & 0.598 & 0.683 & 0.683 \\ 
    \textit{SVM} & \textbf{0.796} & 0.832 & 0.832 & \textbf{0.796} & 0.832 & 0.832 & \textbf{0.636} & \textbf{0.72} & \textbf{0.72} & \textbf{0.621} & \textbf{0.708} & \textbf{0.708} \\ 
    \textit{XGBoost} & 0.784 & 0.788 & 0.788 &  0.785 & 0.788 & 0.788 & 0.604 & 0.684 & 0.684 &  0.591 & 0.673 & 0.673 \\ 
    \textit{FNN (128, 32)} & 0.764 & \textbf{0.848} & \textbf{0.832} &  0.765 & \textbf{0.849} & \textbf{0.833} & 0.6 & 0.692 & 0.696 &  0.594 & 0.687 & 0.689 \\ 
    \hline
    \end{tabular}
    \label{tab:eval1}
\end{table*}

\subsubsection{Evaluation} We evaluate the model performances using the standard metric of accuracy (fraction of correct predictions out of total) on the test set in Table~\ref{tab:eval1}. Given the imbalance in our dataset, we also report the f1-scores (weighted average of class-wise F1 scores). We note that the models are able to achieve a fair degree of accuracy (significantly higher than baselines), and also that using larger embeddings (RoBERTa and DistilBERT) results in better accuracy than smaller USE embeddings, although between RoBERTa and DistilBERT, there is no significant difference in performance. 
We fine-tune the hidden layer parameter for the FNN model and find the combination of using two layers of sizes 128 and 32 as the best configuration. Interestingly, a simple model (SVM) performs at par with a sophisticated model (FNN) as noted from the accuracies in Table~\ref{tab:eval1}. 
\section{Conclusion and Future Work}
\label{sec:conclusion}

We present \emph{ReAct}, a novel and carefully annotated dataset for identifying the nature of textual review comments. These comments play an important role in the evolution of many types of documents. We propose a new taxonomy for fine-grained comment classification in a review scenario. We analyze the properties of the dataset along with some baseline systems for the two identified text classification tasks and analyze their performance. 
Since the two tasks are not completely independent from each other, a multitask learning approach seems desirable. 
The small fraction of labeled data out of a large pool of unlabelled data also calls for a self-supervised learning algorithm using less data.

\bibliographystyle{splncs04}
\bibliography{references}

\end{document}